\documentclass{article}
\usepackage{spconf,amsmath,graphicx}

\usepackage{amsmath,graphicx}
\usepackage{amssymb}
\usepackage{subfigure}
\usepackage{caption}
\usepackage{comment}
\usepackage[table]{xcolor}

\title{A Regularized Attention Mechanism for Graph Attention Networks}
\name{Uday Shankar Shanthamallu$^{\dagger}$, Jayaraman J, Thiagarajan$^{\ddagger}$\thanks{This work was supported in part by the ASU SenSIP Center, Arizona State University. Portions of this work were performed under the auspices of the U.S. Department of Energy by Lawrence Livermore National Laboratory under Contract DE-AC52-07NA27344.} and Andreas Spanias$^{\dagger}$}
\address{$^\dagger$Arizona State University, $^\ddagger$Lawrence Livermore National Laboratory \\
	Email: \{ushantha@asu.edu, jjayaram@llnl.gov, spanias@asu.edu\}}
\begin{document}
%\ninept
%
\maketitle
\begin{abstract}
Machine learning models that can exploit the inherent structure in data have gained prominence. In particular, there is a surge in deep learning solutions for graph-structured data, due to its wide-spread applicability in several fields. Graph attention networks (GAT), a recent addition to the broad class of feature learning models in graphs, utilizes the attention mechanism to efficiently learn continuous vector representations for semi-supervised learning problems. In this paper, we perform a detailed analysis of GAT models, and present interesting insights into their behavior. In particular, we show that the models are vulnerable to heterogeneous rogue nodes and hence propose novel regularization strategies to improve the robustness of GAT models. Using benchmark datasets, we demonstrate performance improvements on semi-supervised learning, using the proposed robust variant of GAT.
\end{abstract}
\section{Introduction}
\label{sec:intro}

Dealing with relational data is central to a wide-range of applications including social networks \cite{eagle2006reality}, epidemic modeling \cite{simon2011exact}, chemistry \cite{pham2018graph}, medicine, energy distribution, and transportation \cite{henderson2012rolx}. Consequently, machine learning formalisms for graph-structured data \cite{hamilton2017representation, latouche2015graphs} have become prominent, and are regularly being adopted for information extraction and analysis. In particular, graph neural networks \cite{scarselli2009graph, niepert2016learning, zhang2018} form an important class of approaches, and they have produced unprecedented success in supervised and semi-supervised learning problems. Broadly, these methods generalize convolutional networks to the case of arbitrary graphs \cite{kipf2016semi, defferrard2016convolutional}, through spectral analysis (e.g. Laplacian \cite{henaff2015deep}) or neighborhood based techniques~\cite{duvenaud2015convolutional}. Graph Attention Networks (GAT)\cite{velickovic2017graph} are the recent addition to this class of methods and they rely solely on attention mechanisms  for feature learning. In contrast to spectral approaches, attention models do not require the construction of an explicit Laplacian operator and can be readily applied to non-Euclidean data. Further, GATs are highly effective, thanks to the recent advances in attention modeling \cite{vaswani2017attention}, and easily scalable. Given the wide-spread adoption of attention models in language modeling and computer vision, it is crucial to understand the functioning and robustness of the attention mechanism.
\begin{figure*}[t]
	\centering
	\subfigure[Cora: Node ID-188, degree-3]{\includegraphics[width=.3\linewidth]{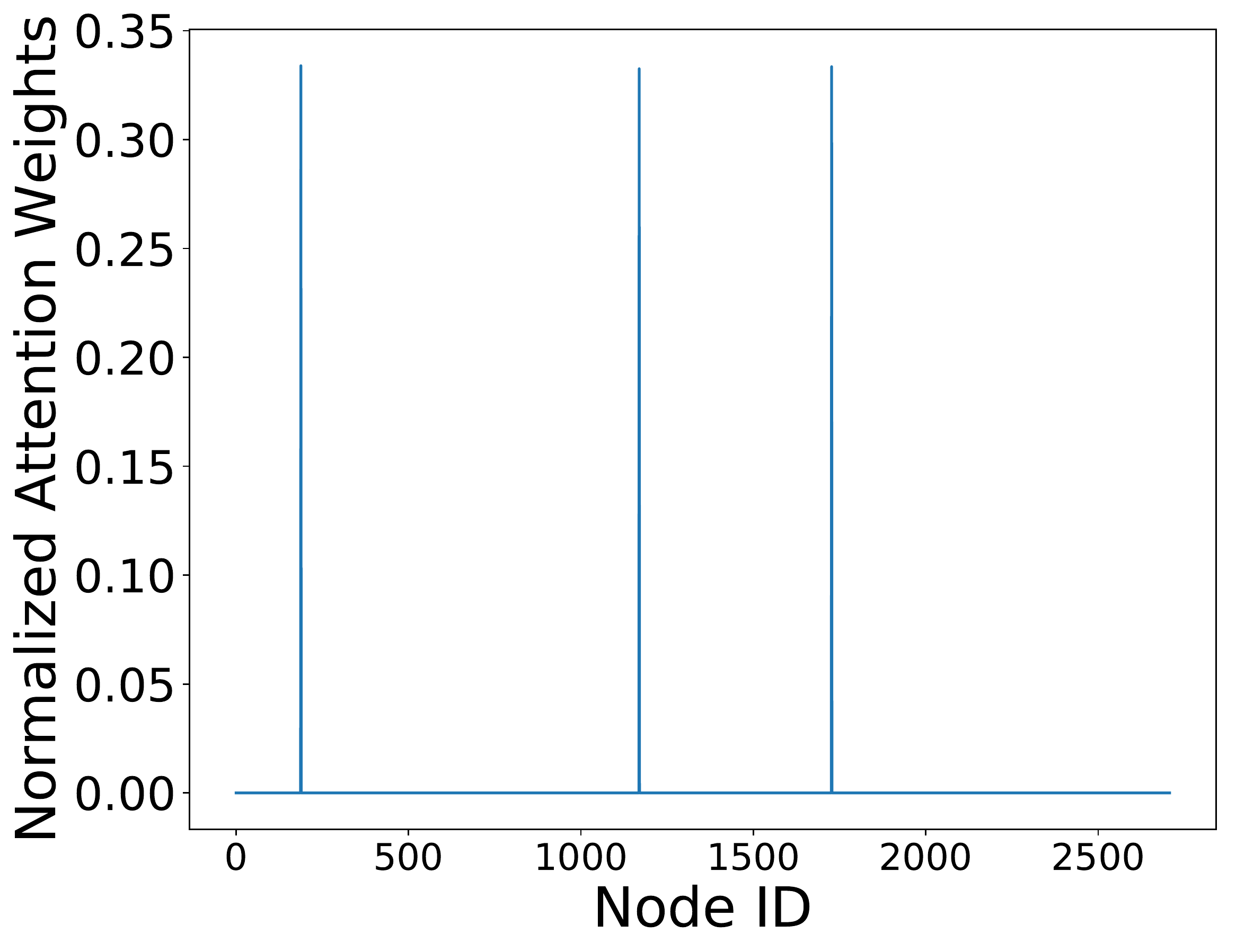}}
	\hspace{0.5in}
	\subfigure[Cora: Node ID-1701, degree-75]{\includegraphics[width=.3\linewidth]{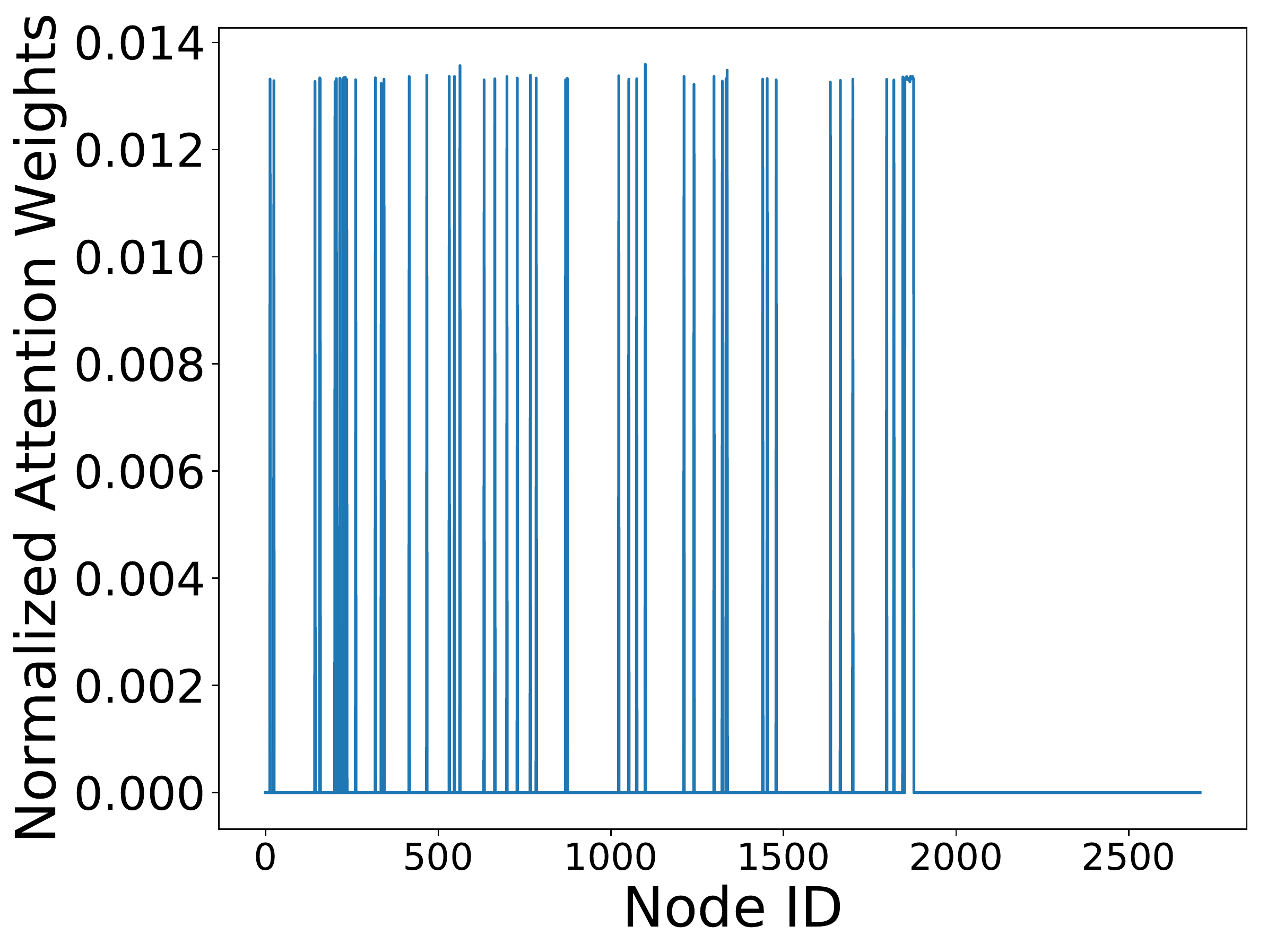}}
	
	\subfigure[Citeseer: Node ID-1591, degree-7]{\includegraphics[width=.3\linewidth]{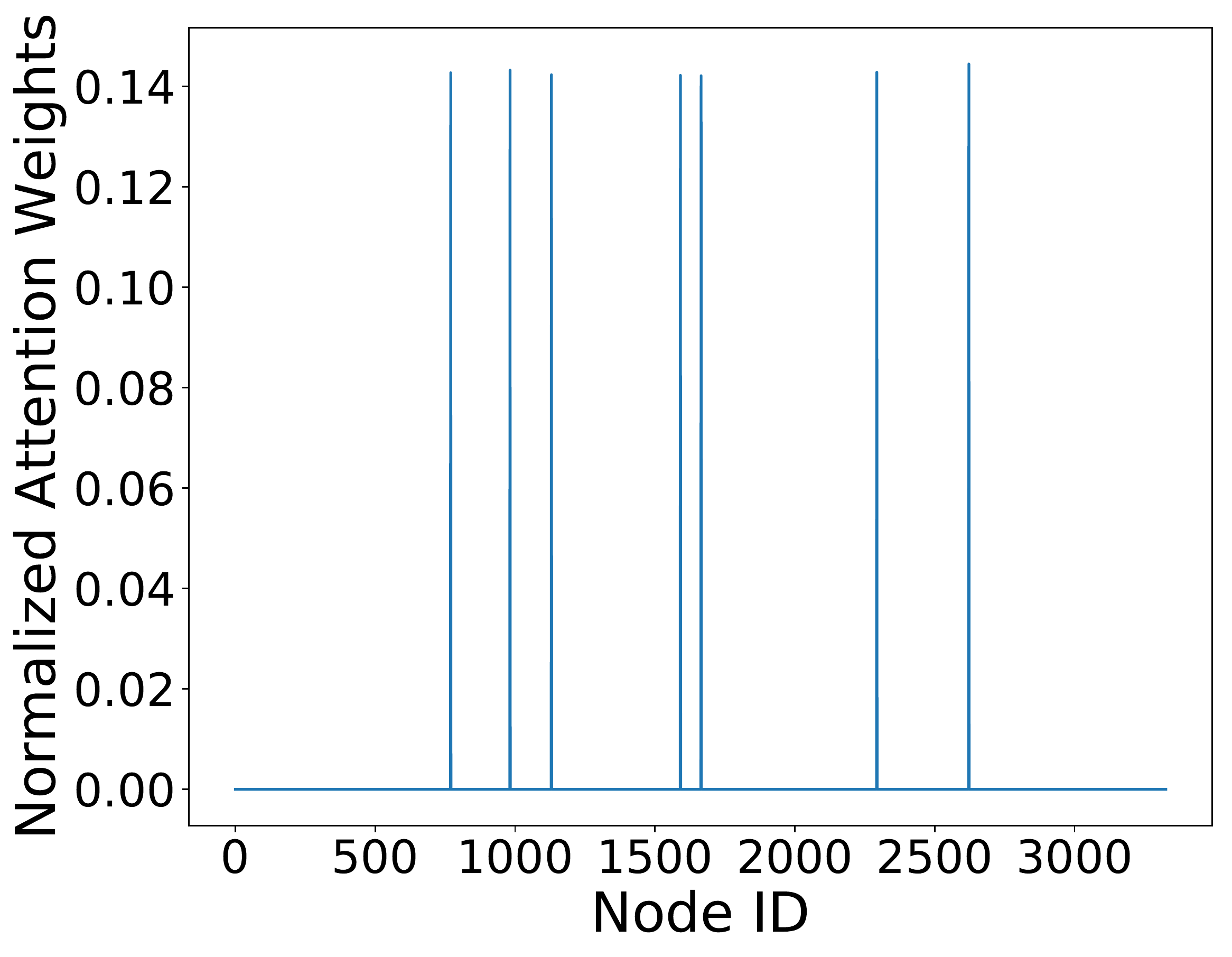}}
	\hspace{0.3in}
	\subfigure[Citeseer: Node ID-582, degree-52]{\includegraphics[width=.3\linewidth]{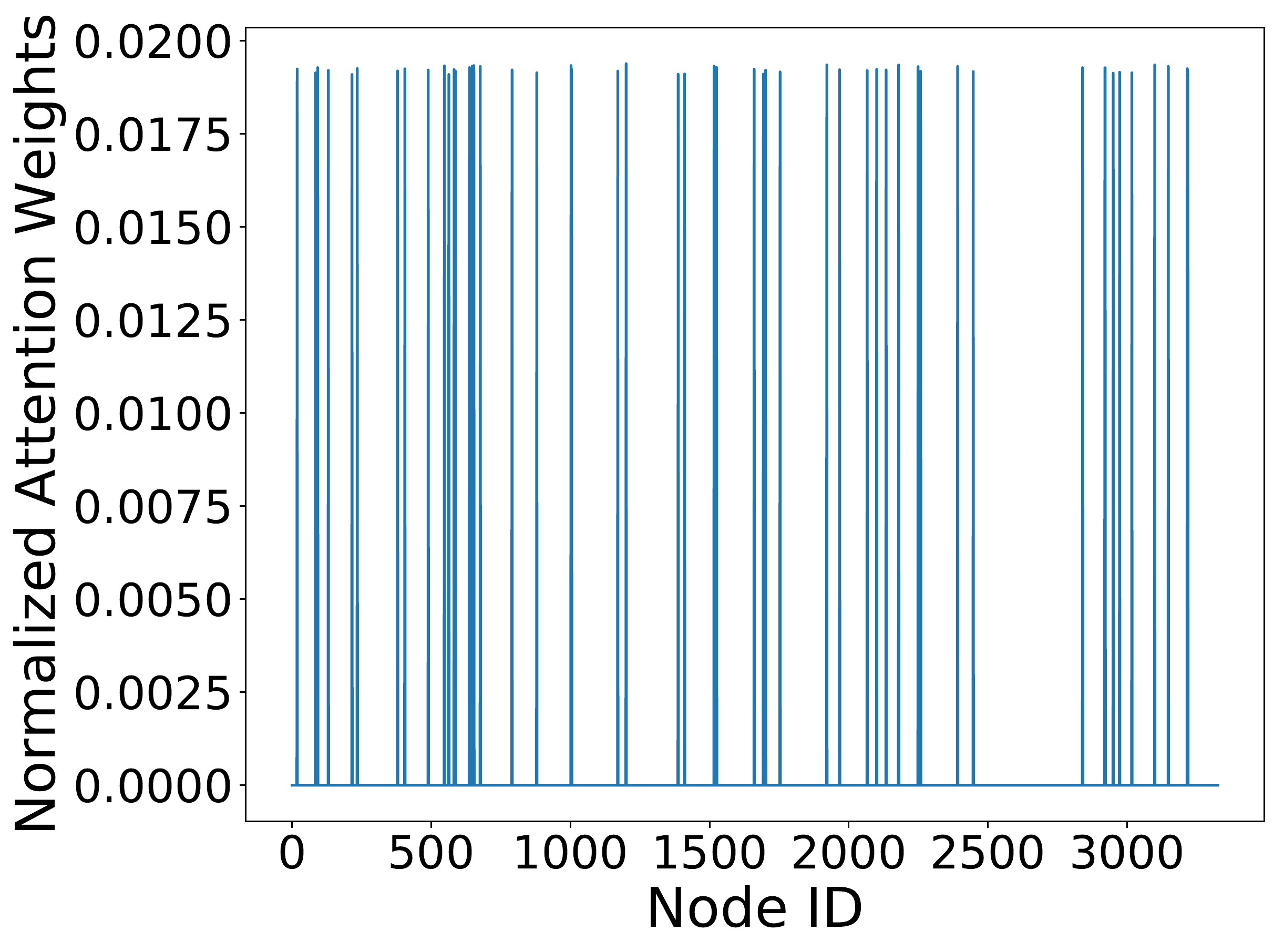}}
	
	\caption{Attention weights learned by the GAT model for Cora and Citeseer datasets. There is a tendency to produce uniform attention weights even when a node has a large neighborhood size.}
	\vspace{-0.1in}
	\label{fig:attention}
\end{figure*} 

In this paper, we perform a detailed analysis of the GAT models, and present interesting insights to their behavior. We regularize the attention mechanism in GAT in order to improve the robustness of the attention models. An attention model parameterizes the local dependencies to determine the most relevant parts of the neighborhood to focus on, while computing the features for a node. Using empirical analysis with GAT, we make an interesting observation that with unweighted graphs, there is a tendency to produce uniform attention weights to all connected neighbors at every node. In other words, all neighbors are chosen with the same importance; consequently nodes with high degree or high valency end up being highly influential to the feature learning. Subsequently, inferencing can be greatly affected by introducing even a small number of "rogue" or "noisy" nodes with high degree into the network structure. For example, in a social network, an entity (node) with an ill intent, can corrupt the network structure by establishing connections with several other nodes  even though it does not necessarily share coherent community association. In practice, such noisy nodes can arise due to measurement errors, availability of partial information while constructing the relational database, or the presence of adversaries specifically designed to make inference challenging. This motivates the need to regularize the attention mechanism and improve robustness of attention models on graphs. 

We propose an improved variant of GAT that analyzes the distribution of attention coefficients, and attempts to minimize global influence of each node and the tendency to produce uniform attention weights across a neighborhood. We achieve this through the inclusion of sparsity based regularization strategies. Using experiments with benchmark network datasets, we demonstrate improvements over standard GAT in semi-supervised learning, thus effectively combating structural noise in graphs.

%Node 1701 in Cora dataset has 75 neibhoring nodes and all of them are given equal importance

\section{Graph Attention Networks}
\label{sec:gat}
We represent an undirected and unweighted graph using the tuple set
$\mathcal{G} = (\mathcal{V}, \mathcal{E})$, where $\mathcal{V}$ denotes the set of nodes with cardinality $|\mathcal{V}|$ = $N$ , $\mathcal{E}$ denotes the set of edges. Each node $v_i$ is endowed with a $d$-dimensional node attribute vector (also referred as the graph signal), $\mathbf{x}_i \in \mathbb{R}^d$. For a given node $v_i$, its closed neighborhood $\mathcal{N}_c(v_i)$ is given by $\{v_i \cup v_j  | e(v_i,v_j) \in \mathcal{E}\}$.

An attention head is the most basic unit in GAT \cite{velickovic2017graph}. A head basically learns a hidden representation for each node by performing a weighted combination of node attributes in the closed neighborhood, where the weights are trainable. In our setup, we consider a simple dot-product attention, similar to the \textit{Transformer} architecture \cite{vaswani2017attention}. Formally, an attention head is comprised of the following steps:

\noindent \textbf{Step 1:} Feed-forward layer that transforms each 
$\mathbf{x}_i \in \mathbb{R}^d  \text{ into }  \tilde{\mathbf{x}}_i  \in \mathbb{R}^{d'}$.

\noindent \textbf{Step 2:} A shared trainable dot-product attention mechanism which learns coefficients for each valid edge in the graph. This is carried out using attributes of the connected neighbors, $ e_{ij} = \langle {Att}, \tilde{\mathbf{x}}_i || \tilde{\mathbf{x}}_j \rangle $, where ${Att}\in \mathbb{R}^{2d'}$ denotes the parameters of the attention function, and $||$ represents concatenation of features from nodes $v_i$ and $v_j$ respectively. 

\noindent \textbf{Step 3:} A softmax layer for normalizing the learned attention coefficients across the closed neighborhood, $ a_{ij} = softmax(e_{ij} ; \forall j \in \mathcal{N}_c(v_i)), \text{ s.t. } 
\sum_{j}^{}a_{ij} = 1$. For simplicity, we represent the normalized attention coefficients for the entire graph as $\mathbf{A} \in \mathbb{R}^{N\times N}$, where $A_{ij}$ denotes the importance of node $v_j$ features in approximating the feature for node $v_i$.

\noindent \textbf{Step 4:} A linear combiner that performs weighted combination of node features with the learned attention coefficients followed by a non-linearity: $\tilde{\mathbf{z}}_i = \sigma(\mathbf{z}_i)$, where $\mathbf{z}_i =  \sum_{j \in \mathcal{N}_c(v_i)} a_{ij}  \tilde{\mathbf{x}}_j$.

\section{Analysis of Attentions Inferred by GAT}
\label{sec:analysis}
%\begin{itemize}
%	\item Show a layout of the inferred attention, like the original GAT paper
%	\item Define the metrics - IQR for each the cases and count (define this as the $\ell_0$ or the $\ell_1$ norm)
%	\item Show this distribution across heads and demonstrate the tendency to produce uniform distributions and over-usage of nodes with high degree
%\end{itemize}

In this section, we perform a detailed analysis of the attention coefficients learned by GAT. The presented observations will motivate the need for strategies to regularize the attention mechanism in attention models. In contrary to our expectation that an attention head might assign different levels of importance to the nodes in the closed neighborhood, our analysis shows that for most of the nodes, the distribution of coefficients in a closed neighborhood is almost always uniform in nature. This is particularly the case when the node degree is low.
\begin{figure*}[t]
	\centering
	\subfigure[Original Cora Dataset without any rogue nodes/edges]{\includegraphics[width=0.4\linewidth]{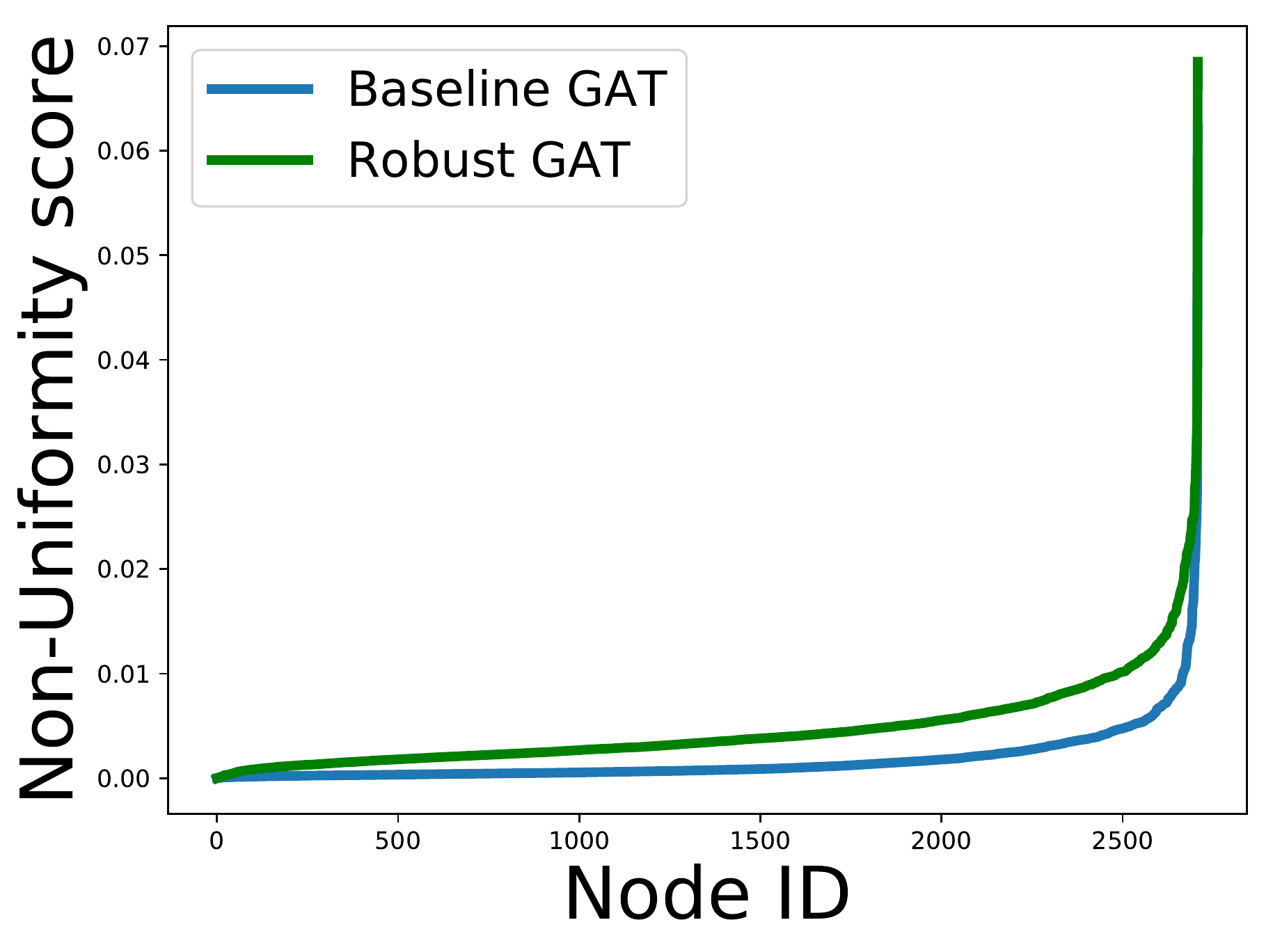}}
	\hspace{0.3in}
	\subfigure[Cora Dataset with 50 rogue nodes and 100 noisy edges per rogue node]{\includegraphics[width=0.4\linewidth]{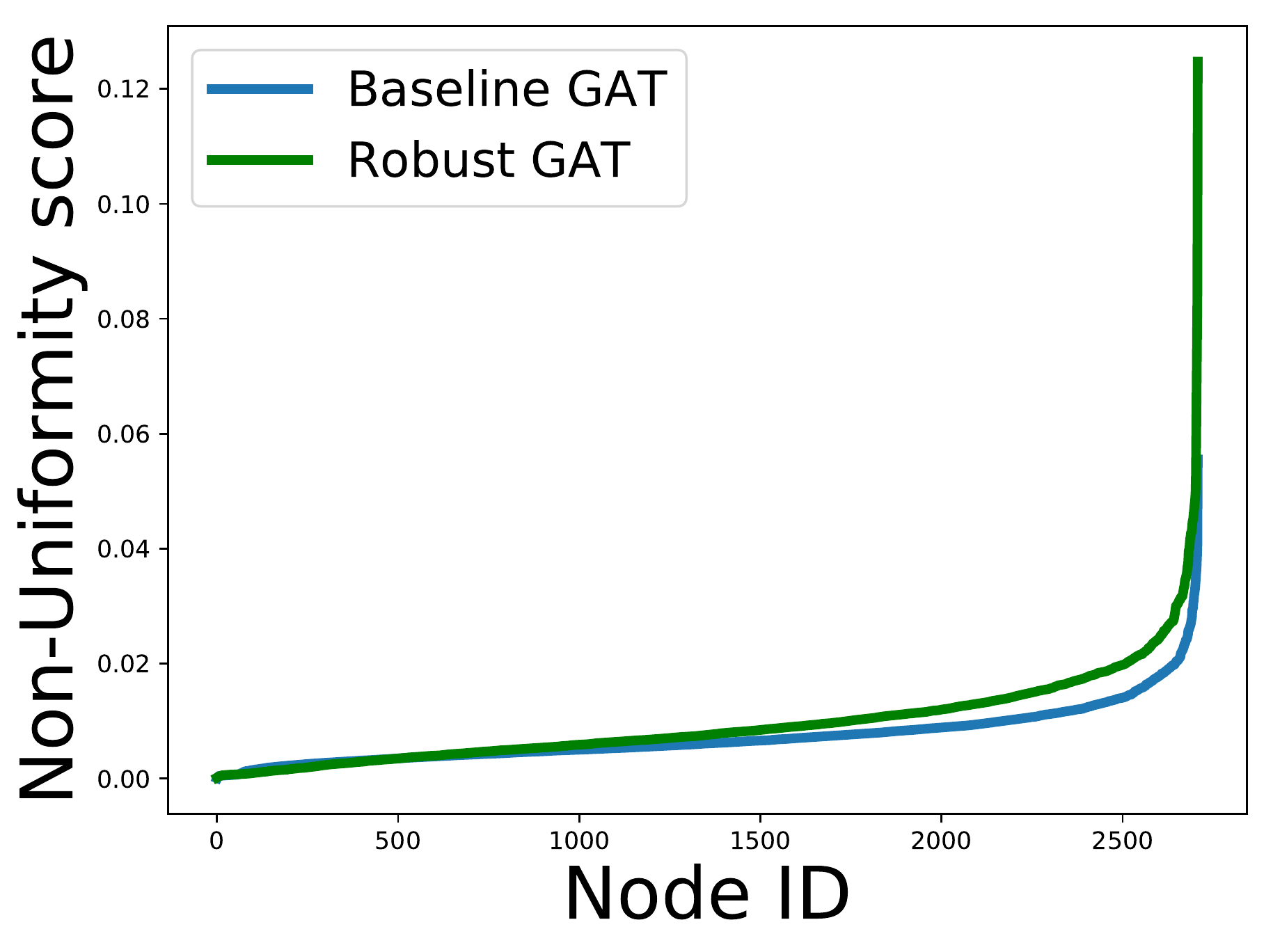}}
	\caption{A plot of non-uniformity scores (sorted) vs node id. Observe that the robust GAT with attention regularization produces higher non-uniformity score indicating it produces non-uniform attention scores across closed-neighborhood.}
	\label{fig:discrepancy}
\end{figure*}

For this study, we consider two benchmark citation datasets, namely \textit{Cora} and \textit{Citeseer} \cite{sen2008collective} (details can be found in Section \ref{sec:results}). Figure \ref{fig:attention} illustrates the weight distribution for two nodes from the Cora and Citeseer dataset with varying degrees. Interestingly, we find that, even for nodes with high degree, the attention mechanism fails to prioritize nodes in a closed neighborhood, and learns uniform weights for all nodes. This is particularly undesirable since the resulting features can be noisy for small neighborhoods, while rogue nodes with large degree can have a much higher impact than expected.
%Although this behavior is reasonable for small neighborhood sizes, we observe this trend even when $|\mathcal{N}_c| = 75$.

In order to quantitatively test our hypothesis on the lack of meaningful structure in the computed attention scores in a closed neighborhood, we use a discrepancy metric given by
\begin{equation}
\label{eq:discrepancy}
	d_i = \frac{\| A_{i,:} - \mathit{U}_i\|_1}{degree(v_i)}
\end{equation}

where $\mathit{U}_i$ is the uniform distribution score for the node $v_i$. So $d_i$ gives a measure of non-uniformity in the learned attention score, a lower discrepancy value indicates a strong uniformity in the attention score and vice versa. For every node, we measure the discrepancy score. Figure \ref{fig:discrepancy} shows a plot of discrepancy score.

\section{Proposed Approach}
\label{sec:approach}
%\begin{itemize}
%	\item Argument - when a node has a lot of neighbors, no one node is exclusively influential and hence compromising uniformity to ignore a few neighbors is fine. This will implicitly reduce participation of nodes in approximation for other nodes. However, when a node has a few neighbors, enforcing uniformity is critical, since dropping a few neighbors might take away lot of vital information .
%	\item Motivate the need to enable GAT models to trade-off between these two objectives.
%	\item Define the modified objective function
%	\item Describe the losses and write comments about the training (if any)
%\end{itemize}

In this section, we describe the proposed regularization strategy, in order to improve the robustness of GAT models to produce non-uniform attention scores. The proposed solution will help combat structured noise in graph datasets. Intuitively, our approach attempts to improve the reliability of local attention structure by systematically limiting the influence of nodes globally. To this end, we build upon the observations in the previous section, and propose two additional regularization terms with respect to the attention mechanism in the original GAT formulation.
%They regularize taking global structure of graph in to account rather than just local structural information (one step neighborhood). 

As described in Section \ref{sec:analysis}, there is a tendency in GAT models, with unweighted graphs, to produce a uniform weight distribution while training the attention mechanism. For a node with a small closed neighborhood, it is prudent to utilize information from all its neighbors in order to approximate the node's latent representation. Consequently, producing uniform attention weights can be a reasonable in such cases. However, if the closed neighborhood contains a rogue node, i.e. a node that cannot be characterized as part of any coherent community in the graph, uniform attention can lead to severe uncertainties in the local approximation. The first regularization strategy limits the global influence of a rogue node in terms of its participation in the approximation of other nodes's features. We introduce a penalty for \textit{exclusivity}. 
For every node $v_j$, this term is defined as the $\ell_1$-norm of attention coefficients assigned to that node in an attention head: $\|{A}_{:,j}\|_1$. Generalizing this to $K$ independent heads in the GAT model, we obtain
\begin{equation}
\mathrm{L}_{excl} = \frac{1}{NK} \sum_{k=1}^{K} \sum_{j=1}^{N} \sum_{i=1}^{N}  |{A}^k_{ij}|	
\end{equation}This prevents any one node (or subset of nodes) in the graph to be exclusively influential to the overall feature inferencing. In other words, this does not allow a node with high degree to arbitrarily participate in the approximation of all its neighbors. This is particularly important when those nodes are noisy or adversarial

\begin{figure*}[t]
	\centering
	\subfigure[Cora dataset]{\includegraphics[width=0.38\linewidth]{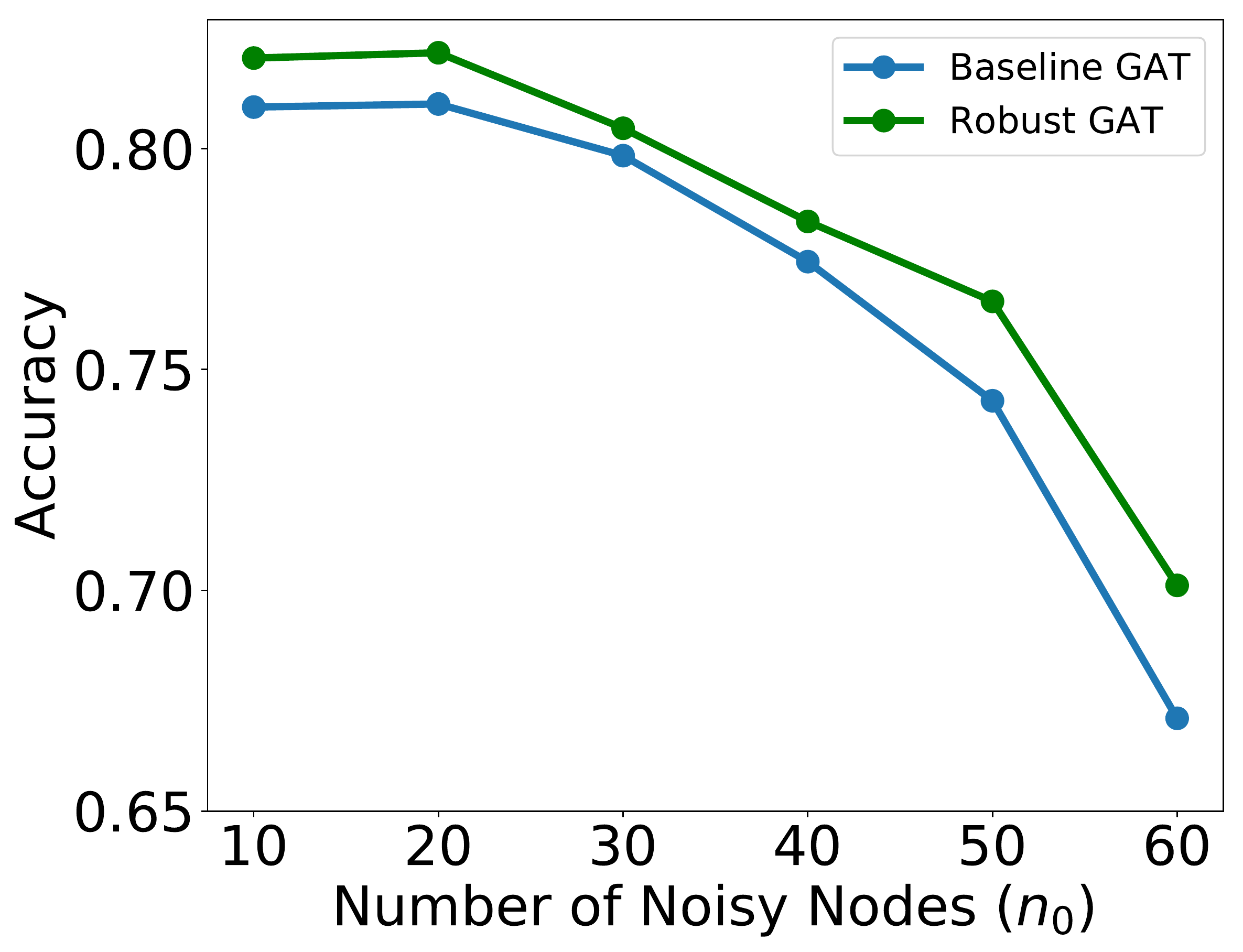}}
	\hspace{0.3in}
	\subfigure[Citeseer dataset]{\includegraphics[width=0.38\linewidth]{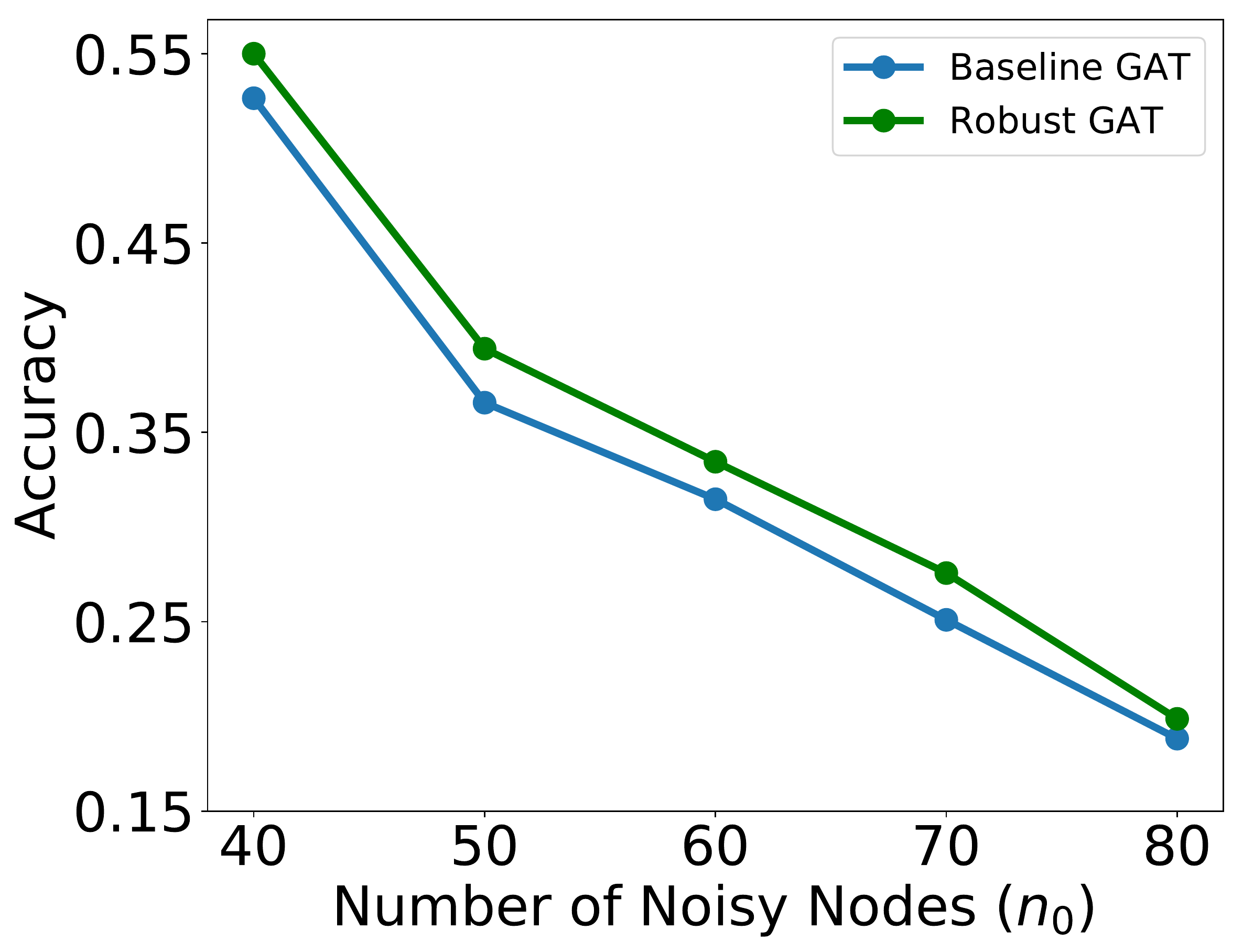}}
	\caption{Semi-supervised learning performance under the presence of noisy nodes. Test accuracies are obtained by aggregating the performance from 20 random trials.}
	\label{fig:results}
\end{figure*}

The second regularization term explicitly penalizes the uniformity in the attention scores. Note that this is similar to the equation (\ref{eq:discrepancy}). However, we found that the below form produces better results.
\begin{equation}
	\mathrm{L}_{nonunif} = \frac{1}{K} \sum_{k=1}^{K} \frac{1}{N} \sum_{i=1}^{N}  \|\mathbf{A}^k_i\|_0 -  \text{deg}(v_i).	
\end{equation}Here, we measure the $\ell_0$ norm, i.e. the number of nodes in the neighborhood that have been assigned a non-zero attention weight. By comparing it to the degree of that node, we penalize if all nodes in the neighborhood participate in the approximation. So we maximize the $\mathrm{L}_{nonunif}$ and $K$ here represents the number of independent heads.

By including the two proposed regularization terms to the original loss function of GAT, we obtain
\begin{equation}
\mathrm{L_{total}} = \mathrm{L}_{GAT} + \lambda_1 \mathrm{L}_{excl}- \lambda_2 \mathrm{L}_{nonunif}  
\end{equation}
where the hyperparameters $\lambda_1$ and $\lambda_2$ are used to weight the penalty terms with respect to the classification loss. Note that, $\mathrm{L}_{GAT}$ includes an additional $\ell_2$ regularization on the learned model parameters. We optimize all the model parameters with respect to $\mathrm{L_{total}}$. We refer to this formulations as the \textit{Robust GAT}. With the proposed modification to the GAT objective, we can now limit the global influence of a node while simultaneously producing a non-uniform attention scores in a closed-neighborhood.
%Real-world relational data when obtained without any pre-processing can be very noisy. Learning and inferencing from a noisy graph is challenging. In scenarios where even few rogue nodes are present, a lot of other nodes end up learning from the rogue nodes and there by corrupting the inference performed on the graph. The proposed loss function does not allow any particular node (or subset of nodes) to exclusively have strong influence on the overall graph structure.

%For nodes with relatively small neighborhood and sparse connections, it still achieves uniformity among the neighborhood and exclusivity loss is less.

\section{Experiment Setup and Results}
\label{sec:results}
%\begin{itemize}
%	\item datasets - Cora and Pubmed
%	\item experiment setup -- varying number of rogue nodes with different degree levels.
%	\item Plot of results
%	\item Discuss impact of hyperparameters
%	\item Show distribution of IQR vs $\ell_0$ for the modified training.
%\end{itemize}
 
In this section, we describe the experimental setup for evaluating the impact of adversarial nodes on the GAT model, and present the results from the proposed robust variant. We use two benchmark citation networks: \textit{Cora} and \textit{Citeseer} \cite{sen2008collective}. In both the datasets, documents are treated as nodes, and citations among the documents are encoded as undirected edges. Additionally, each node (document) is endowed with an attribute vector (bag-of-word representations). We follow the experimental setup for training and inferencing similar to that of GAT \cite{velickovic2017graph} and perform transductive learning. We perturb the graph structure by explicitly introducing rogue nodes which have noisy edges with regular nodes in the graph. This represents a scenario where the presence of adversaries can make the inferencing more challenging. 

We introduce structured noise to  graph datasets in the following manner: First, we sample $n_0$ nodes uniformly at random (without replacement) from the validation set, and delete all the existing edges for each of the selected nodes. We then add a total of $m_0$ arbitrary edges for each of the $n_0$ nodes. Note, the nodes to which the edges were established were also chosen at random, but from the entire graph. We specifically introduced noisy nodes only in the validation set to show the impact of adversaries on the overall performance, even when they are not part of training or testing. For comparison, we generate results from the baseline GAT approach in each of these cases. We mainly compare to GAT as it has the state of the art accuracies on the datasetsFor a fair comparison, our architectures and the hyperparameter choices were fixed to be the same for both \textit{Baseline GAT} and the \textit{Robust GAT} approaches. Note that, \textit{Robust GAT} has two additional hyperparameters, $\lambda_1$ and $\lambda_2$ which were set to $0.01$ and $0.5$ respectively. 

For \textit{Cora} dataset, we vary the number of noisy nodes $n_0$ from $10$ to $60$ and fix the number of noisy edges per node at $m_0 = 100$. Since the order of the \textit{Citeseer} dataset is high, we vary $n_0$ from $40$ to $80$ and set $m_0=500$. For each case of $n_0$, we performed $20$ independent realizations and report the average. The results from this case study are shown in Figure \ref{fig:results}. 
%We observe that, when the number of noisy edges are fixed at a high value as $500$, even $40$ or $50$ noisy nodes results in severe performance degradation. For example, with the cora dataset, when $n_0$ is increased from $60$ to $70$, the average accuracy of GAT on the test set decreases by nearly $50\%$. In comparison, the proposed robust variant produces improvements in range of $3\%$ to $12\%$, depending on the number of noisy nodes. However, when the number of noisy nodes is increased beyond a certain point, the additional regularizations are not sufficient to combat the large amounts of noise. 

\section{Conclusion}
\label{sec:conclusion}
In this paper, we analyzed the attention mechanism in graph attention networks, and showed that they are highly vulnerable to noisy nodes with high degrees in the graph. This can be attributed to the surprising behavior of GAT in producing uniform attention to all connected neighbors at every node. In order to alleviate this limitation, we proposed a robust variant of GAT, that minimizes the global influence of a node (or a subset of nodes) and also produces non-uniform attention scores for a closed neighborhood. Using benchmark datasets, we demonstrated improvements in the semi-supervised learning performance in the presence of structural noise.

% We showed that the GAT model has a tendency to produce uniform attention weights  for all connected nodes in the given neighborhood. Also, we proposed two additional losses: non-uniformity loss and exclusivity loss; when added to the existing GAT architecture, it greatly improves the robustness of the attention mechanism for each attention head. 

% Below is an example of how to insert images. Delete the ``\vspace'' line,
% uncomment the preceding line ``\centerline...'' and replace ``imageX.ps''
% with a suitable PostScript file name.
% -------------------------------------------------------------------------

%\vfill
%\pagebreak

\bibliographystyle{IEEEbib}
\bibliography{refs}

\end{document}